\documentclass[runningheads]{llncs}

\usepackage{eccv}

\usepackage{eccvabbrv}

\usepackage{graphicx}
\usepackage{booktabs}
\newcommand{\includegraphicsmaybe}[2][]{%
\IfFileExists{#2}{\includegraphics[#1]{#2}}{\fbox{\begin{minipage}[c][0.22\textheight][c]{0.95\linewidth}\centering Missing file:\\\texttt{\detokenize{#2}}\end{minipage}}}%
}

\usepackage{ifxetex}
\ifxetex\else\usepackage[accsupp]{axessibility}\fi
\usepackage{hyperref}

\usepackage{orcidlink}

\usepackage{pgfplots} %引用包

\usepackage{tikz}%导入Tikz宏包
\usepackage{xcolor}
\definecolor{blue}{RGB}{60,132,196}
\definecolor{red}{RGB}{207,78,56}
\definecolor{gray}{RGB}{146,146,161}
\definecolor{green4}{RGB}{46, 139, 87}
\definecolor{red2}{RGB}{149,9,30}
\usepackage{multirow} 
\usepackage{colortbl}
\definecolor{lightgray}{gray}{0.9}
\usepackage{hhline} % For double horizontal lines
\usepackage{pdflscape}

\usepackage{amssymb}
\usepackage{amsfonts}
\usepackage{amsmath}
\spnewtheorem{assumption}{Assumption}{\bfseries}{\itshape}
\usepackage{graphicx}
\usepackage{pifont}
\usepackage{extpfeil}

\usepackage{dsfont}

\usepackage{svg}
\usepackage{mathrsfs}
\DeclareMathAlphabet{\mathpzc}{OT1}{pzc}{m}{it}
\definecolor{pink1}{RGB}{239 41 140}
\definecolor{pink}{RGB}{30,144,255}
\definecolor{BarrierColor}{RGB}{25, 162, 30}
\definecolor{BicycleColor}{RGB}{79, 241, 199}
\definecolor{BusColor}{RGB}{198, 1, 90}
\definecolor{CarColor}{RGB}{150, 232, 149}
\definecolor{ConstVehColor}{RGB}{147, 83, 113}
\definecolor{MotorcycleColor}{RGB}{143, 43, 251}
\definecolor{PedestrianColor}{RGB}{9, 41, 148}
\definecolor{TrafficConeColor}{RGB}{86, 155, 103}
\definecolor{TrailerColor}{RGB}{118, 122, 211}
\definecolor{TruckColor}{RGB}{46, 6, 61}
\definecolor{DrivableSurfColor}{RGB}{185, 217, 16}
\definecolor{OtherFlatColor}{RGB}{142, 44, 241}
\definecolor{SidewalkColor}{RGB}{242, 218, 92}
\definecolor{TerrainColor}{RGB}{225, 12, 72}
\definecolor{ManmadeColor}{RGB}{111, 209, 144}
\definecolor{VegetationColor}{RGB}{182, 34, 111}

\providecommand{\ie}{\textit{\ie}}
\providecommand{\eg}{\textit{\eg}}

\pgfplotsset{compat=1.18}

\usepackage{marvosym}
\usepackage{ifsym}
\begin{document}

\title{FDR-Occ: Factorized Dense Routing for Full-Spectrum 3D Occupancy Prediction} 

\titlerunning{FDR-Occ}

% Author list (camera-ready). Corresponding author marked with envelope symbol.
\author{Dubing Chen\inst{1} \and
Huan Zheng\inst{1} \and
Tianyi Yan\inst{1} \and
Yucheng Zhou\inst{1} \and \\
Runzhou Tao\inst{2} \and
Zhongying Qiu\inst{2} \and
Jianfei Yang\inst{3} \and
Jianbing Shen\inst{1,}{\textsuperscript{\Letter}}}

% Abbreviated list of authors for the running head.
\authorrunning{D.~Chen et al.}

% Institution list.
\institute{SKL-IOTSC, CIS, University of Macau \and
Chongqing Afari Intelligent Drive Co., Ltd. \and
Geely Automobile Research Institute (Ningbo) Co., Ltd.}

\maketitle

% !TEX root = main.tex
\begin{abstract}
Vision-based 3D occupancy prediction fundamentally relies on the 2D-to-3D view transformation. Current paradigms predominantly utilize explicit physical projection, which artificially restricts the routing matrix to strict, sparse camera rays. While computationally efficient, this imposes a severe \textit{Locality Bottleneck}, preventing the network from constructing holistic contextual understanding and degrading sharply when camera extrinsics are unreliable or absent. To break this bottleneck, we abstract view transformation as unconstrained bipartite routing and propose \textbf{Factorized Dense Routing (FDR)}. By approximating dense 2D-to-3D mixing through hierarchical tensor contractions, FDR guarantees a fully-global receptive field with tractable, sub-quadratic complexity. Crucially, the mandatory spatial contraction in dense routing exposes a fundamental \textit{Resolution-Context Trade-off}. To address this, we introduce a \textbf{Resolution-Context Decoupled Architecture}. We factorize the 3D space into a global macroscopic topological anchor (via FDR) and precise local geometric planes (via explicit projection). This decoupling enables global semantic inference and exact surface localization to complement each other without mutual compromise. Extensive experiments demonstrate that our framework achieves state-of-the-art performance on the Occ3D-nuScenes and Occ3D-Waymo benchmarks. More notably, in an uncalibrated setting where physical extrinsics are withheld, our global routing internalizes the implicit multi-camera rig topology and exhibits substantially stronger structural robustness than physical-projection baselines under the same protocol.

\end{abstract}

\begin{figure}[t]
    \centering
    \includegraphics[width=1.\linewidth]{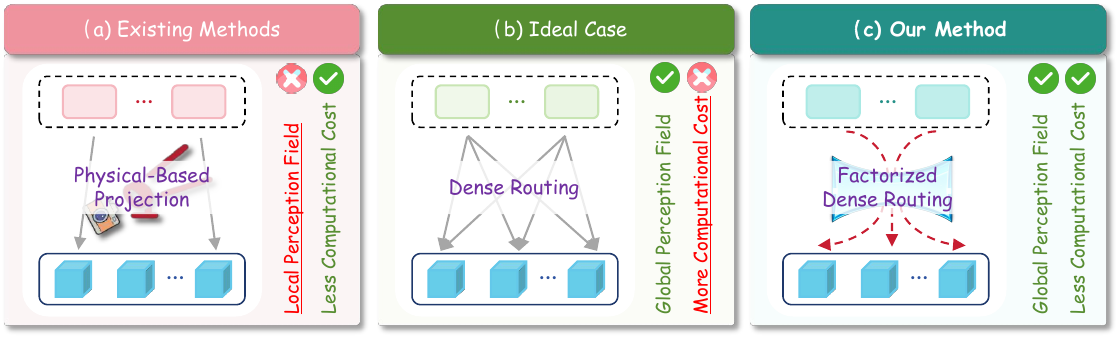}
    \caption{\textbf{Evolution of View Transformation (VT) Routing.} \textbf{Left:} Explicit physical projection~\cite{philion2020lift,chen2025alocc} enforces a rigid ray-wise sparsity, causing structural fragmentation. \textbf{Middle:} Ideal dense routing enables $100\%$ reachability but suffers from an intractable quadratic complexity. \textbf{Right:} Our \textbf{Factorized Dense Routing (FDR)} structurally factorizes the dense mapping into hierarchical tensor contractions. It achieves fully-global reachability with tractable complexity, while effectively internalizing holistic 3D topology that is unreachable via localized physical rays.}
    \label{fig:motivation}
\end{figure}

\section{Introduction}
\label{sec:intro}

\let\thefootnote\relax\footnotetext{\Letter \ Corresponding author: \textit{Jianbing Shen}. This work was supported by the National Natural Science Foundation of China (No. 624B2002), the Science and Technology Development Fund of Macau SAR (FDCT) under grants 0134/2025/RIA2 and 0102/2023/RIA2, and the Jiangyin Hi-tech Industrial Development Zone under the Taihu Innovation Scheme (EF2025-00003-SKL-IOTSC).}

Vision-based 3D occupancy prediction has emerged as a fundamental task for autonomous driving and robotic navigation \cite{huang2023tri,wei2023surroundocc,tian2024occ3d,wang2023openoccupancy,li2025uniscene,yan2025drivingsphere,yang2025xscene,li2025omninwm}. The core challenge lies in the 2D-to-3D view transformation (VT), which governs how perspective image features are lifted into a 3D volumetric space. Historically, the field has been dominated by methods relying on explicit physical ray priors. These can be broadly categorized into active forward-projection paradigms, represented by Lift-Splat-Shoot (LSS) family \cite{philion2020lift,li2023bevstereo,huang2022bevdet4d,chen2025alocc} and reactive backward-querying mechanisms, leveraging 3D-to-2D cross-attention \cite{li2022bevformer,li2023fb}.

Despite their widespread adoption, physical-prior-based methods are bottlenecked by a fundamental architectural flaw: \textbf{the strict ray-wise localization of receptive fields}. In these frameworks, the 2D-to-3D routing matrix is artificially constrained to an extreme sparsity pattern tightly aligned with physical lines-of-sight \cite{li2023fb}. This rigid constraint causes \textit{Restricted Reachability} during VT: a 2D pixel can only broadcast its features strictly along its physical ray (as shown in Figure~\ref{fig:motivation}), leaving massive 3D volumetric regions completely unreachable to direct image evidence. Consequently, the network fails to construct global topological understanding, splatting continuous structures (\eg, holistic road layouts or heavily occluded cross-camera vehicles) as isolated fragments. Furthermore, it relies on an overly rigid assumption: the availability of perfectly precise camera parameters. When physical calibrations are noisy, perturbed, or entirely absent, this localized projection mathematically collapses.

To break this locality bottleneck, we argue that the VT operator must structurally evolve from a locally constrained physical ray toward fully global reachability. Mathematically, as we formally prove in Section \ref{sec:preliminary}, the function family defined by local physical routing is merely a strictly constrained subset of the unconstrained global routing family. Thus, a global routing operator possesses a strictly higher representational upper bound. However, solving the global routing involves explicitly computing a dense bipartite mixing matrix across massive spatial dimensions, which incurs an intractable quadratic complexity. By analogy with the architectural evolution from fully-connected networks to factorized convolutional fields, we propose \textbf{Factorized Dense Routing (FDR)}. FDR introduces a scalable structural inductive bias, approximating the dense lifting operator with a hierarchy of localized tensor operations. By progressively contracting the 2D spatial dimensions and expanding into the 3D volume, FDR guarantees a fully-global receptive field while reducing the computational burden by nearly two orders of magnitude.

While FDR theoretically breaks the bottleneck of localized routing, realizing this linear efficiency practically necessitates progressive spatial aggregation. This mathematical property exposes a fundamental \textbf{Resolution-Context Trade-off} in view transformation: expanding the global receptive field inherently entails aggregating fine-grained spatial details. Consequently, FDR excels at abstracting macroscopic global topology but prioritizes holistic scene layouts over microscopic geometric precision. To exploit unconstrained global context without sacrificing pixel-exact spatial precision, we propose a \textbf{Resolution-Context Decoupled Architecture}. 
We intentionally factorize the 3D representation space into two orthogonal pathways. We configure FDR as the \textit{Global Context Pathway}. Concurrently, we position the classical LSS as the \textit{Local Resolution Pathway}; by circumventing spatial contraction entirely, it acts as an optimal extractor to carve precise local geometries. These two orthogonal capacities are unified on the shared Bird's-Eye-View (BEV) plane, resolving the trade-off between unconstrained global semantic inference and precise geometric boundary preservation.

We extensively evaluate our method on the Waymo Open \cite{sun2020waymo} and the nuScenes datasets \cite{caesar2020nuscenes}, establishing new state-of-the-art results for 3D occupancy prediction under the ResNet-50 backbone with minimal temporal frames. Crucially, we design a rigorous \textbf{uncalibrated setting} where explicit camera parameters are entirely withheld. In such scenarios where classical methods degrade sharply, our method learns to internalize the implicit rig topology, demonstrating robustness to unconstrained real-world conditions.

Our main contributions are summarized as follows:
\begin{itemize}
    \item We formally identify the \textit{Restricted Reachability} bottleneck inherent in explicit physical projection and characterize its structural limitation in constructing unconstrained global context.
    \item We propose Factorized Dense Routing, a novel global routing view transformation operator. By approximating dense 2D-to-3D mixing through hierarchical spatial tensor contractions, FDR achieves a fully-global receptive field with tractable complexity.
    \item To resolve the fundamental \textit{Resolution-Context Trade-off}, we design a decoupled volumetric architecture. By combining a globally-connected Holistic Context Anchor (via FDR) with high-resolution geometric planes (via LSS), our framework captures both unconstrained macroscopic scene layouts and precise local geometric boundaries.
    \item Our method achieves state-of-the-art performance on the Waymo Open and the nuScenes datasets. Experimental results under a strictly uncalibrated configuration demonstrate that it can robustly infer implicit multi-camera topology even in the complete absence of explicit camera parameters.
\end{itemize}

\section{Related work}
\label{sec:literature}
\subsection{3D Semantic Occupancy Prediction}
3D semantic occupancy prediction~\cite{tian2024occ3d,wei2023surroundocc,wang2023openoccupancy} aims to infer dense volumetric occupancy and semantics of the 3D scene from inputs such as images or point clouds. Early work focused on monocular Semantic Scene Completion (SSC)~\cite{roldao20223dssc,liu2018seessc,song2017semanticssc,xia2023scpnetssc}. MonoScene~\cite{cao2022monoscene} pioneered monocular 3D semantic completion from a single RGB image, while subsequent architectures such as VoxFormer~\cite{li2023voxformer} and OccFormer~\cite{zhang2023occformer} densified sparse voxel queries and addressed class imbalance through dual-path transformer designs.

More recent autonomous-driving research has shifted toward multi-camera joint perception, predicting the occupancy and semantics of the 3D voxel grid surrounding the ego-vehicle. Occ3D~\cite{tian2024occ3d}, SurroundOcc~\cite{wei2023surroundocc}, and OpenOccupancy~\cite{wang2023openoccupancy} established surround-view occupancy benchmarks on nuScenes~\cite{caesar2020nuscenes} and catalyzed this direction. Follow-up work then advanced along several axes. To explore compact intermediate representations, TPVFormer~\cite{huang2023tri}, SparseOcc~\cite{liu2024fully}, OPUS~\cite{wang2024opus}, and GaussianFormer~\cite{huang2024gaussianformer} move beyond dense voxels. To improve efficiency, FastOcc~\cite{hou2024fastocc}, FlashOcc~\cite{yu2024panoptic}, and ALOcc~\cite{chen2025alocc} adopt height-compression strategies that accelerate computation. To reduce reliance on 3D supervision, RenderOcc~\cite{pan2024renderocc}, SelfOcc~\cite{huang2024selfocc}, OccNeRF~\cite{zhang2023occnerf}, GaussTR~\cite{jiang2024gausstr}, and SurfelOcc~\cite{wang2026surfelocc} learn occupancy from 2D supervision alone. To exploit temporal cues, GDFusion~\cite{chen2025rethinking} and GaussianWorld~\cite{zuo2025gaussianworld} study history fusion, while ALOcc~\cite{chen2025alocc}, ViewFormer~\cite{li2024viewformer}, OccWorld~\cite{zheng2024occworld}, OccSora~\cite{wang2024occsora}, and DrivingSphere~\cite{yan2025drivingsphere} target occupancy flow or 4D occupancy prediction. A further line incorporates auxiliary inputs such as LiDAR, radar, or satellite imagery~\cite{wolters2025unleashing,chen2025alocc,ming2024occfusion,chen2025saocc}.

\subsection{2D-to-3D View Transformation}

The transformation of 2D image features into a unified 3D volumetric space is the core module of vision-based 3D occupancy prediction. Classical designs inherit from early 3D object detection~\cite{huang2021bevdet,liu2022petr,li2022bevformer} and fall into two paradigms: top-down and bottom-up transformations. Top-down transformer-based methods attach learnable queries to the 3D structure and aggregate 2D spatio-temporal features into them, as in BEVFormer~\cite{li2022bevformer}, VoxFormer~\cite{li2023voxformer}, PanoOcc~\cite{wang2024panoocc}, and SparseOcc~\cite{liu2024fully}. Such methods are highly learnable and can adapt the 2D-to-3D geometric mapping from data. Bottom-up methods instead rely on explicit depth estimation to project multi-view features into the 3D or BEV space. This paradigm was pioneered by LSS~\cite{philion2020lift} and matured in the BEVDet~\cite{huang2021bevdet}, BEVDepth~\cite{li2022bevdepth}, and BEVStereo~\cite{li2023bevstereo} line of detectors. It better exploits explicit physical structure and often attains stronger performance, but suffers from feature sparsity~\cite{li2023fb} and difficulty in handling occlusion~\cite{chen2025alocc}.

Several methods combine the two paradigms: FB-Occ~\cite{li2023fbocc} and COTR~\cite{ma2023cotr} obtain LSS-initialized 3D features and then refine them with query-based attention. Others enhance the learnability and occlusion handling of the LSS pipeline: ALOcc~\cite{chen2025alocc} introduces a probabilistic transfer of surface-depth probabilities into occluded regions; CausalOcc~\cite{chen2025semantic} integrates the separate camera-calibration, coordinate-mapping, and geometry-prediction modules into a single end-to-end learnable module with a semantics-supervised geometry loss; DHD~\cite{wu2025deep} introduces height priors into the view transformation; and GDFusion~\cite{chen2025rethinking} performs the 2D-to-3D transformation with history-fused geometry. Nevertheless, all these methods depend heavily on camera parameters for physical projection, and their aggregation still operates within ray-aligned neighborhoods. When calibration is noisy or unavailable, the projection degrades and cross-ray context remains hard to establish. In contrast, we propose a globally-reachable 2D-to-3D view transformation that both supplements physical projection with global context and enables camera-parameter-free occupancy prediction.

\section{Methodology}
\label{sec:method}

\subsection{Problem Formulation}
\label{sec:formulation}
The core objective of vision-based 3D occupancy prediction is to transform multi-view perspective images into a dense, semantically annotated 3D volumetric grid. Given a set of surround-view images, a 2D backbone first extracts dense image-plane features $\mathcal{F}_{2D} \in \mathbb{R}^{N \times C \times H \times W}$, where $N$ is the number of camera views, $C$ is the feature dimension, and $H, W$ are the spatial resolutions. 
The crux of the pipeline is the View Transformation module, an operator $\mathcal{T}$ that lifts the perspective features into a 3D ego-car coordinate space to form a volumetric tensor $\mathcal{V}_{3D} = \mathcal{T}(\mathcal{F}_{2D}; \Theta) \in \mathbb{R}^{C \times X \times Y \times Z}$. Here, $\Theta$ denotes the provided camera calibration parameters (intrinsics and extrinsics), and $X, Y, Z$ represent the spatial dimensions of the 3D grid. Following the VT module, the 3D encoder and the decoder processes $\mathcal{V}_{3D}$ to output the final voxel-wise semantic probabilities $\mathcal{O} \in \mathbb{R}^{B \times X \times Y \times Z}$, where $B$ is the number of semantic categories. The representational capacity of the view transformation operator $\mathcal{T}$ is therefore the primary bottleneck for the entire system.

\begin{figure}[t]
    \centering
    \includegraphics[width=\linewidth]{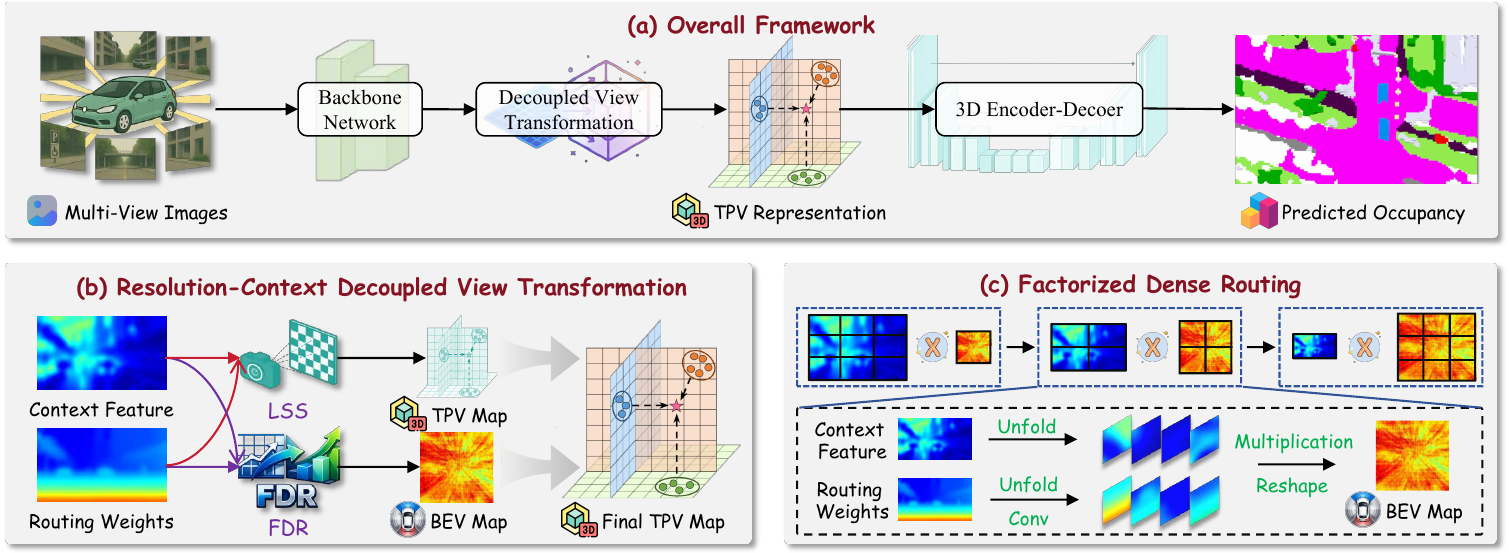}
    \caption{\textbf{Overview of our Resolution-Context Decoupled Architecture.} \textbf{(a) Overall Framework.} \textbf{(b) Decoupled 2D-to-3D Projection:} To solve the spatial-contextual trade-off, the view transformation is factorized into two orthogonal pathways. Explicit LSS preserves high-frequency local geometry via orthogonal planes (TPV Map), while FDR captures unconstrained global topology via a Z-compressed BEV anchor. \textbf{(c) Factorized Dense Routing:} FDR approximates intractable dense global routing via hierarchical tensor contractions.}
    \label{fig:framework}
\end{figure}
\subsection{View Transformation as Spatial Routing: The Locality Bottleneck}
\label{sec:preliminary}

To systematically analyze the representational capacity of existing VT operators, we abstract the 2D-to-3D lifting process into a generalized spatial routing framework~\cite{chen2025alocc}. This perspective allows us to mathematically expose the inherent locality bottlenecks of physical projection. 

Let $S_0 = H W$ and $V_T = X Y Z$ denote the total spatial elements in the 2D multi-view images and the 3D ego-space, respectively. The view transformation can be universally formulated as a bipartite routing:
\begin{equation}
    \text{vec}(\mathcal{V}_{3D}) = \mathbf{W} \, \text{vec}(\mathcal{F}_{2D})
\end{equation}
where $\mathbf{W} \in \mathbb{R}^{V_T \times S_0}$ represents the routing matrix governing the information flow from perspective pixels to 3D voxels, and $\mathrm{vec}(\cdot)$ defines the flattened feature. 

\vspace{0.5ex}\noindent\textbf{Explicit Physical Projection.} Practically computing a dense unconstrained $\mathbf{W}$ incurs a prohibitive quadratic complexity of $\mathcal{O}(S_0 V_T)$. To make the 2D-to-3D lifting computationally tractable, classical physical-prior-based methods~\cite{philion2020lift} artificially enforce extreme sparsity on the routing matrix. Specifically, LSS predicts a categorical depth distribution over $D$ discrete bins along the camera ray of each pixel, and splats the probability-weighted features strictly into those localized bins.

We can formally characterize this computational compromise as a ray-local support constraint. Let $\mathbf{W}[:, s] \in \mathbb{R}^{V_T}$ denote the $s$-th column of the routing matrix, representing the projection weights from a single 2D pixel $s$. Mathematically, the \textit{support set} of this vector, denoted as $\text{supp}(\mathbf{W}[:, s])$, is defined as the set of 3D voxel indices where the routing weight is strictly non-zero (\ie, the specific voxels that successfully receive direct evidence from pixel $s$). 

Given the calibration $\Theta$, let $\mathcal{R}_\Theta(s) \subseteq \{1, \dots, V_T\}$ denote the small subset of 3D voxels that physically intersect with the geometric ray of token $s$. The classical LSS explicitly limits the non-zero elements of $\mathbf{W}$ to this ray, defining a highly constrained feasible set:
\begin{equation}
    \mathcal{S}_{ray}(\Theta) = \Big\{ \mathbf{W} : \text{supp}(\mathbf{W}[:, s]) \subseteq \mathcal{R}_\Theta(s), \ \forall s \Big\}
\end{equation}

\vspace{0.5ex}\noindent\textbf{The Defect: Restricted Reachability.} By restricting the 2D-to-3D routing matrix to a highly sparse pattern strictly aligned with the ray support set, explicit lifting secures computational efficiency but suffers from severe locality limitations. This rigid structural constraint leaves a massive number of 3D voxels completely unreachable or only weakly reachable to direct image evidence during the VT stage. Consequently, when observing continuous macroscopic topologies (\eg, cross-camera vehicles or occlusion boundaries), the network independently splats them as isolated structural fragments. 

Because the necessary cross-ray geometric constraints are mathematically blocked by $\mathcal{R}_\Theta(s)$, LSS-based architectures are forced to rely on post-hoc 3D/BEV convolutional networks to progressively diffuse information and bridge these fragments. This \textit{posterior remedy} is computationally inefficient, prone to generating fractured occupancy predictions, and structurally suboptimal. 

To overcome this inherent suboptimality, the VT operator must theoretically evolve from the locally constrained space toward a fully global reachability. Let $\mathcal{S}_{global} = \mathbb{R}^{V_T \times S_0}$ denote the completely unconstrained dense routing space, where a 2D pixel can theoretically route to any 3D voxel. For any feasible set $\mathcal{S}$, let $\mathcal{F}(\mathcal{S}) = \{ f_{\mathbf{W}}: \mathcal{F}_{2D} \mapsto \mathbf{W}\mathcal{F}_{2D} \mid \mathbf{W} \in \mathcal{S} \}$ denote the induced function family. We establish the following theoretical motivation:

\vspace{0.5ex}\noindent\textbf{Proposition 1 (Expressivity Inclusion by Relaxing Support).} \textit{For any given camera calibration $\Theta$, the ray-local routing is mathematically merely a subset constraint bounded by the physical geometry. Therefore, the feasible set satisfies $\mathcal{S}_{ray}(\Theta) \subset \mathcal{S}_{global}$, and the induced function families mathematically satisfy $\mathcal{F}(\mathcal{S}_{ray}(\Theta)) \subseteq \mathcal{F}(\mathcal{S}_{global})$, with the inclusion being strict whenever cross-ray voxel correlations carry non-trivial information.}

This proposition shows that relaxing the routing constraints yields a function family with representational capacity at least as large as, and generically strictly larger than, that of the ray-constrained family. We therefore propose to approximate global dense routing at tractable complexity.

\subsection{Factorized Dense Routing (FDR)}
\label{sec:fdr}

While Proposition 1 establishes the theoretical necessity of $\mathcal{S}_{global}$, explicitly parameterizing a fully unconstrained routing matrix incurs an intractable computational complexity of $\mathcal{O}(S_0 V_T C)$. To practically instantiate unconstrained global reachability, we propose \textbf{Factorized Dense Routing (FDR)}, as shown in Figure~\ref{fig:framework}~(c). Motivated by the architectural evolution from fully-connected dense networks to hierarchical convolutional fields, FDR approximates the intractable dense matrix $\mathbf{W}$ as a product of $T$ localized, stage-wise tensor operations:
\begin{equation}
    \mathbf{W} \approx \mathbf{M}^{(T)} \mathbf{M}^{(T-1)} \dots \mathbf{M}^{(1)}
\end{equation}
where each $\mathbf{M}^{(t)}$ acts as a sparse block-matrix operator performing local dense routing between a contracted 2D patch and an expanded 3D micro-volume.

\vspace{0.5ex}\noindent\textbf{Stage-wise Spatial Contraction and Expansion.} Let $\mathcal{X}^{(t)} \in \mathbb{R}^{S_t \times V_t \times C}$ denote the flattened spatial tensor at stage $t$. The algorithm is initialized with $\mathcal{X}^{(0)} = \mathcal{F}_{2D} \in \mathbb{R}^{S_0 \times 1 \times C}$, where the initial 3D volume dimension is degenerated at $V_0 = 1$. Our objective is to progressively contract the 2D spatial size $S_t \to 1$ while exponentially expanding the 3D volume size $V_t$ toward the macroscopic 3D dimensions. Between adjacent stages, standard 2D convolutions on the 2D space are applied to refine the spatial features.

At each stage $t$, we partition the 2D spatial grid into non-overlapping patches with adaptive sizes $P_t = \Delta_h^{(t)} \times \Delta_w^{(t)}$. We unfold the features within each patch, yielding:
\begin{equation}
    \tilde{\mathcal{X}}^{(t-1)} = \text{Unfold}_{P_t}(\mathcal{X}^{(t-1)}) \in \mathbb{R}^{S_t \times V_{t-1} \times C \times P_t}
\end{equation}
where $S_t = S_{t-1}/P_t$, with adaptive zero-padding applied to ensure exact divisibility. 
Subsequently, each 2D patch is dense-routed to an expanded 3D micro-volume containing $K_t = \Delta_x^{(t)} \times \Delta_y^{(t)} \times \Delta_z^{(t)}$ tokens via a dynamically generated routing matrix $\mathbf{A}^{(t)} \in \mathbb{R}^{S_t \times V_{t-1} \times P_t \times K_t}$. This local dense routing is executed as a tensor contraction over the patch dimension $P_t$:
\begin{equation}
    \mathcal{Y}^{(t)} = \tilde{\mathcal{X}}^{(t-1)} \otimes \mathbf{A}^{(t)}, \quad \mathcal{Y}^{(t)} \in \mathbb{R}^{S_t \times V_{t-1} \times C \times K_t},
\end{equation}
where $\otimes$ denotes batched matrix multiplication along the $P_t$ axis. Finally, a $\text{Fold}$ operation merges the newly expanded $K_t$ dimension into the 3D volume axis, yielding the updated spatial tensor $\mathcal{X}^{(t)} \in \mathbb{R}^{S_t \times V_t \times C}$, with $V_t = V_{t-1} \times K_t$. The exact progression of these spatial dimensions is detailed in Table \ref{tab:stage_shape}.

\begin{table}[t]
\caption{\textbf{Progression of Spatial Dimensions across FDR Stages.} To maintain linear complexity, the 2D dimensions progressively contract via $\Delta_h, \Delta_w$, while the 3D dimensions exponentially expand via $\Delta_x, \Delta_y, \Delta_z$.}
\label{tab:stage_shape}
\centering
\begin{tabular}{l@{\hspace{4mm}}c@{\hspace{4mm}}c@{\hspace{4mm}}c@{\hspace{4mm}}c}
\toprule
\textbf{Stage $t$} & \textbf{2D Resolution} & \textbf{3D Volume} & \textbf{Patch $P_t$} & \textbf{Expansion $K_t$} \\
\midrule
$0$ (Input) & $H \times W$ & $1 \times 1 \times 1$ & - & - \\
$1$ & $\frac{H}{\Delta_h^{(1)}} \times \frac{W}{\Delta_w^{(1)}}$ & $\Delta_x^{(1)} \times \Delta_y^{(1)} \times \Delta_z^{(1)}$ & $\Delta_h^{(1)} \Delta_w^{(1)}$ & $\Delta_x^{(1)} \Delta_y^{(1)} \Delta_z^{(1)}$ \\
$\dots$ & $\dots$ & $\dots$ & $\dots$ & $\dots$ \\
$T$ (Output) & $1 \times 1$ & $X \times Y \times Z$ & - & - \\
\bottomrule
\end{tabular}
\end{table}

\vspace{0.5ex}\noindent\textbf{Adaptive Geometric Context Routing.} The dynamic weights $\mathbf{A}^{(t)}$ dictate the critical probability mapping. Instead of utilizing rigid static parameters, we dynamically generate them. Specifically, we fuse the intermediate spatial features with an aligned geometric context $\mathcal{E}_{geo}^{(t)}$, process them through a lightweight mapping network $g_\theta^{(t)}$ (comprising standard convolutions), and subsequently unfold the resulting pre-routing logits:
\begin{equation}
    \mathbf{A}^{(t)} = \sigma\left( \text{Unfold}_{P_t}\Big( g_\theta^{(t)}\big( \mathcal{X}^{(t-1)} \parallel \mathcal{E}_{geo}^{(t)} \big) \Big) \right)
\end{equation}
where $\parallel$ denotes channel concatenation, and $\sigma$ is the Softmax function normalized over the expanded $K_t$ micro-volume dimension. 

Crucially, $\mathcal{E}_{geo}^{(t)}$ comprises explicit local depth features (derived from a lightweight depth head) and Plücker ray coordinates $\mathcal{E}_{ray} = \text{MLP}\left( \mathbf{d} \parallel (\mathbf{o} \times \mathbf{d}) \right)$. Conditioning the routing weights on $\mathcal{E}_{geo}^{(t)}$ treats physical geometry as a soft inductive bias rather than a rigid structural constraint. Unlike LSS, which hard-zeros routing weights outside the physical ray, this context-aware conditioning allows FDR to discover implicit 3D correlations.

\vspace{0.5ex}\noindent\textbf{Complexity and Theoretical Advantage.} 
While fully unconstrained global routing dictates an intractable complexity of $\mathcal{C}_{global} = \mathcal{O}(S_0 V_T C)$, FDR factorizes this process to reduce the relative computational cost to a strict ratio of $\mathcal{C}_{FDR} / \mathcal{C}_{global} = \mathcal{O} \big( \sum_{t=1}^T \frac{S_{t-1}}{S_0} \frac{V_t}{V_T} \big)$. 
Physically, early stages are computationally trivialized by the unexpanded 3D volume ($V_t \ll V_T$), while late stages are exponentially accelerated by the highly contracted 2D grid ($S_{t-1} \ll S_0$). 
As mathematically detailed in the Supplementary Material, under our practical settings ($T=3$), FDR achieves fully-global reachability while requiring merely $\sim 1.4\%$ of the dense theoretical computation.

FDR also subsumes classical LSS as a special case. As proved in the Supplementary Material, \textit{the classical LSS operator is strictly a degenerate instance of FDR} (obtained when $T=1$, $P_1 = 1 \times 1$, and weights are heavily masked). This inclusion ($\mathcal{F}_{LSS} \subseteq \mathcal{F}_{FDR}$) guarantees that FDR's hypothesis space strictly contains that of explicit physical priors, allowing it to internalize multi-camera rig topologies even when physical extrinsics are uncalibrated or absent.

\subsection{Resolution-Context Decoupled Unification}
\label{sec:unification}

As established above, circumventing the intractable quadratic complexity of dense global routing intrinsically requires progressively contracting the spatial dimensions via $P_t > 1$. This mathematical necessity reveals a fundamental \textbf{Resolution-Context Trade-off}: significantly reducing the computational cost necessitates spatial coarsening. Therefore, a single monolithic routing operator cannot simultaneously maintain pixel-exact 2D spatial precision and achieve fully-global contextual reachability. 

Rather than forcing a single operator to violate this structural bound, we propose a decoupled architecture that factorizes the 3D representation space into two orthogonal pathways~(Figure~\ref{fig:framework}~(b)):
\begin{itemize}
    \item \textbf{Global Context Pathway ($P_t > 1$):} The mandatory patch-wise aggregation in FDR explicitly expands the contextual footprint. It excels at gathering wide-range global semantics (\eg, continuous road layouts and holistic occlusion relationships) to achieve the unconstrained reachability of $\mathcal{S}_{global}$.
    \item \textbf{Local Resolution Pathway ($P_t = 1 \times 1$):} Explicit LSS strictly avoids any cross-patch aggregation. While this Kronecker-delta-like routing bounds its receptive field to zero ($\mathcal{S}_{ray}$), it perfectly preserves exact 2D spatial precision, acting as an optimal pathway to carve sharp local geometric boundaries.
\end{itemize}

To unify these orthogonal capacities without the memory overhead of dense 3D fusion, we design a volumetric decomposition. We instantiate the local pathway via classical LSS to yield a high-resolution precise surface volume $\mathcal{V}_{LSS} \in \mathbb{R}^{C \times X \times Y \times Z}$. Concurrently, rather than forcing the resolution-coarsened FDR to reconstruct sharp 3D surfaces, we configure its final stage to compress the Z-axis. This outputs a globally connected, gravity-aligned \textit{Holistic Context Anchor} $\mathcal{V}_{global} \in \mathbb{R}^{C \times X \times Y \times 1}$. 

We then orthogonally pool the high-resolution $\mathcal{V}_{LSS}$ into three axis-aligned planes ($\mathcal{P}_{xy}, \mathcal{P}_{xz}, \mathcal{P}_{yz}$). Because global driving topologies inherently reside in the BEV plane, the global-local unification is performed via addition on the shared XY plane:
\begin{equation}
    \tilde{\mathcal{P}}_{xy} = \mathcal{P}_{xy} + \mathcal{V}_{global}
\end{equation}

This formulation combines the unbounded contextual reachability of dense routing with the local geometric precision of sparse physical priors. 

\subsection{Occupancy Decoding and Training}

The unified plane $\tilde{\mathcal{P}}_{xy}$ is processed by a dedicated 2D convolutional encoder, while the remaining geometric planes $\mathcal{P}_{xz}$ and $\mathcal{P}_{yz}$ are processed by a secondary shared 2D encoder. To reconstruct the dense 3D volume, $\tilde{\mathcal{P}}_{xy}$ is first passed through a linear projector to recover the Z-axis dimension from its channels. The encoded $\mathcal{P}_{xz}$ and $\mathcal{P}_{yz}$ planes are then spatially broadcasted along their respective missing axes. Finally, these three expanded tensors are element-wise summed and fed into an MLP decoder to predict the voxel-wise semantic probabilities.
The framework is optimized end-to-end using a joint multi-task objective:
\begin{equation}
    \mathcal{L} =  \mathcal{L}_{BCE} + \mathcal{L}_{dice} +  \mathcal{L}_{depth} +  \mathcal{L}_{sem}, 
\end{equation}
where the 3D occupancy loss combines Binary Cross-Entropy (BCE) and Dice loss to address voxel-class imbalance. For auxiliary supervision, $\mathcal{L}_{depth}$ explicitly guides the depth distribution in the LSS branch, and $\mathcal{L}_{sem}$ enforces 2D perspective-view segmentation following ALOcc \cite{chen2025alocc}.

\section{Experiments}
\label{sec:experiments}
\subsection{Experimental Setup}
\label{sec:setup}

\vspace{0.5ex}\noindent\textbf{Datasets.} We evaluate the proposed decoupled architecture on two standard autonomous driving benchmarks: Occ3D-nuScenes \cite{tian2024occ3d} and Occ3D-Waymo \cite{tian2024occ3d}. 
The Occ3D-nuScenes dataset, derived from the nuScenes dataset \cite{caesar2020nuscenes}, provides dense 3D occupancy annotations encompassing 17 foreground semantic classes. It contains 600 training sequences and 150 validation sequences, with a predefined voxel resolution of $0.4\text{m} \times 0.4\text{m} \times 0.4\text{m}$ covering the spatial range of $[-40\text{m}, 40\text{m}] \times [-40\text{m}, 40\text{m}] \times [-1\text{m}, 5.4\text{m}]$. 
Conversely, the Occ3D-Waymo dataset is constructed from the Waymo Open Dataset \cite{sun2020waymo}. It offers an expanded scale of 798 training and 202 validation sequences with 14 foreground semantic categories, presenting a more complex and expansive urban driving environment.

\vspace{0.5ex}\noindent\textbf{Evaluation Metrics.} Following the established evaluation protocols \cite{tian2024occ3d}, we adopt the mean Intersection over Union (mIoU) metric to quantify the geometric and semantic accuracy of the 3D occupancy predictions. We also report the geometric-only Intersection over Union (IoU) to isolate and assess pure structural completeness.

\vspace{0.5ex}\noindent\textbf{Implementation Details.} 
For a fair comparison with existing baselines, we utilize ResNet-50 \cite{resnet} as the 2D image backbone. The input image resolutions are set to $256 \times 704$ for nuScenes and $640 \times 960$ for Waymo. For our Factorized Dense Routing (FDR), we employ a 3-stage configuration ($T=3$) where, since the Z-axis is compressed to $1$, the BEV micro-volumes are expanded by area factors of $10\times10$, $5\times5$, and $4\times4$, respectively. Correspondingly, the 2D spatial dimensions are contracted with patch sizes $P_t$ of $4 \times 4$, $2 \times 4$, and $2 \times 2$. The model is optimized end-to-end using AdamW \cite{zhou2024towardsadam} with a learning rate of $2 \times 10^{-4}$ and a global batch size of 16 for a total of 24 training epochs. Our experiments on Occ3D-nuScenes focus on settings with minimal temporal frames to explicitly emphasize the structural improvements derived from our 2D-to-3D View Transformation architecture. Due to the scarcity of single-frame baselines on Occ3D-Waymo, we follow ALOcc \cite{chen2025alocc} to fuse 16 temporal frames when evaluating on this dataset.

\begin{table*}[t]
    \centering
    \caption{\textbf{3D Semantic Occupancy Prediction Results on Occ3D-nuScenes Validation Set.} The upper block reports the single-frame setting (no temporal voxel feature fusion); the lower block reports the setting with one historical frame, where the suffix \textbf{+1f} denotes the incorporation of one historical frame. The top performance within each block is highlighted in \textbf{bold}.}
    \renewcommand{\arraystretch}{1.1}
    \resizebox{1.\textwidth}{!}{
    \begin{tabular}{l|c|c|cc|ccccccccccccccccc}
    \toprule
    Method & Backbone & Size & \rotatebox{90}{\textbf{mIoU}} & \rotatebox{90}{\textbf{IoU}}
    & \rotatebox{90}{others} 
    & \rotatebox{90}{barrier} 
    & \rotatebox{90}{bicycle} 
    & \rotatebox{90}{bus} 
    & \rotatebox{90}{car} 
    & \rotatebox{90}{cons. veh.} 
    & \rotatebox{90}{motor.} 
    & \rotatebox{90}{pedes.} 
    & \rotatebox{90}{tfc. cone} 
    & \rotatebox{90}{trailer} 
    & \rotatebox{90}{truck} 
    & \rotatebox{90}{drv. surf.} 
    & \rotatebox{90}{other flat} 
    & \rotatebox{90}{sidewalk} 
    & \rotatebox{90}{terrain} 
    & \rotatebox{90}{manmade} 
    & \rotatebox{90}{vegetation} \\ 
    \midrule
    FB-Occ \cite{li2023fbocc} & ResNet-50 & $256 \times 704$ & 35.7 & 66.5 & 11.2 & 40.0 & 23.2 & 43.1 & 46.6 & 21.8 & 23.3 & 25.9 & 23.5 & 29.5 & 33.7 & 77.6 & 39.0 & 47.3 & 51.0 & 37.5 & 33.0 \\
    DHD-S \cite{wu2025deep} & ResNet-50 & $256 \times 704$ & 36.5 & - & 10.6 & 43.2 & 23.0 & 40.6 & 47.3 & 21.7 & 23.3 & 23.9 & 23.4 & 31.8 & 34.2 & 80.2 & 41.3 & 50.0 & 54.1 & 38.7 & 33.5 \\
    COTR \cite{ma2023cotr} & ResNet-50 & $256 \times 704$ & 39.1 & 69.6 & - & - & - & - & - & - & - & - & - & - & - & - & - & - & - & - & - \\
    BEVDetOcc \cite{huang2022bevdet4d} & ResNet-50 & $256 \times 704$ & 37.1 & 70.4 & 9.2 & 45.6 & 15.9 & 42.1 & 50.0 & 23.0 & 21.0 & 21.9 & 22.7 & 31.1 & 36.8 & 80.2 & 38.7 & 51.6 & 55.2 & 45.4 & 40.2 \\
    LightOcc-S \cite{zhang2024lightweight} & ResNet-50 & $256 \times 704$ & 37.9 & - & 11.7 & 45.6 & 25.4 & 43.1 & 48.7 & 21.4 & 25.6 & 26.6 & 29.2 & 33.2 & 35.1 & 80.0 & 41.8 & 50.4 & 53.9 & 39.4 & 34.0 \\
    ProtoOcc \cite{kim2024protoocc} & ResNet-50 & $256 \times 704$ & 39.6 & - & - & - & - & - & - & - & - & - & - & - & - & - & - & - & - & - & - \\
    ALOcc \cite{chen2025alocc} & ResNet-50 & $256 \times 704$ & 40.1 & 70.2 & 12.4 & 46.4 & 26.2 & 42.6 & 52.1 & 25.6 & 27.5 & 28.1 & 30.2 & 32.9 & 39.5 & 81.2 & 42.5 & 52.6 & 56.0 & 46.1 & 39.7 \\
    CausalOcc \cite{chen2025semantic} & ResNet-50 & $256 \times 704$ & 40.9 & 70.7 & 13.3 & 47.3 & 28.5 & 44.0 & 52.9 & 25.3 & 28.9 & 29.2 & 31.9 & 35.6 & 39.3 & 81.3 & 43.2 & 52.8 & 55.9 & 46.3 & 40.6  \\
    \midrule
    \rowcolor{pink!10} \textbf{Ours} & ResNet-50 & $256 \times 704$ & \textbf{41.2} & \textbf{71.0} & 13.6 & 46.5 & 28.3 & 43.4 & 53.0 & 26.8 & 29.1 & 29.6 & 31.6 & 36.4 & 39.9 & 81.6 & 44.3 & 53.0 & 55.5 & 47.2 & 40.6 \\
\midrule
     COTR \cite{ma2023cotr} & ResNet-50 & $256 \times 704$ & 41.4 & - & 12.2 & 48.5 & 29.1 & 44.7 & 53.3 & 27.0 & 29.2 & 28.9 & 31.0 & 35.0 & 39.5 & 81.8 & 42.5 & 53.7 & 56.9 & 48.2 & 42.1 \\
     DHD-M \cite{wu2025deep} & ResNet-50 & $256 \times 704$ &41.5 &-&12.7 &48.7 &26.3 &43.2 &52.9& 27.3& 28.5& 28.5 &30.0 &35.8 &40.2 &83.1 &44.7 &54.7 &57.7 &48.9 &42.1 \\
     SA-Occ \cite{chen2025saocc} & ResNet-50 & $256 \times 704$&40.7 &- &10.9 &48.5 &23.5 &45.8 &52.8 &24.5 &23.6 &24.2 &22.7 &35.0 &40.3 &83.5 &44.1 &55.5 &60.0 &51.2 &45.3 \\
    \rowcolor{pink!10} \textbf{Ours (+1f)} & ResNet-50 & $256 \times 704$ & \textbf{42.2} & \textbf{72.1} & 14.2 & 48.7 & 28.8 & 44.8 & 53.8 & 26.8 & 29.9 & 30.3 & 33.0 & 35.4 & 41.3 & 82.6 & 45.2 & 54.2 & 57.3 & 49.2 & 42.2 \\
    \bottomrule
    \end{tabular}
    }
    \label{tab:occ_cls}
\end{table*}

\begin{table*}[t]
  \centering
  \caption{\textbf{3D Semantic Occupancy Prediction Results on Occ3D-Waymo Validation Set.} The best results are highlighted in \textbf{bold}.}
  \renewcommand{\arraystretch}{1.1}
  \resizebox{1.\textwidth}{!}{
  \begin{tabular}{l|c|*{15}{c}}
   \toprule
    Method & \rotatebox{90}{\textbf{mIoU}} & \rotatebox{90}{Gen. Obj.} & \rotatebox{90}{Vehicle} & \rotatebox{90}{Pedestrian} & \rotatebox{90}{Sign} & \rotatebox{90}{Bicyclist} & \rotatebox{90}{Traffic Light} & \rotatebox{90}{Pole} & \rotatebox{90}{Cons. Cone} & \rotatebox{90}{Bicycle} & \rotatebox{90}{Building} & \rotatebox{90}{Vegetation} & \rotatebox{90}{Tree Trunk} & \rotatebox{90}{Road} & \rotatebox{90}{Walkable} 
    \\
    \midrule
    BEVFormer-w/o TSA
    & 23.87 & 7.50 & 34.54 & 21.07 & 9.69 & 20.96 & 11.48 & 11.48 & 14.06 & 14.51 & 23.14 & 21.82 & 8.57 & 78.45 & 56.89 \\
    BEVFormer \cite{li2022bevformer}  & 24.58 & 7.18 & 36.06 & 21.00 & 9.76 & 20.23 & 12.61 & 14.52 & 14.70 & 16.06 & 23.98 & 22.50 & 9.39 & 79.11 & 57.04  \\
    SOLOFusion \cite{park2022time} & 24.73 & 4.97 & 32.45 & 18.28 & 10.33 & 17.14 & 8.07 & 17.83 & 16.23 & 19.30 & 31.49 & 28.98 & 16.93 & 70.95 & 53.28 \\
    BEVFormer-WrapConcat & 25.07 & 6.20 & 36.17 & 20.95 & 9.56 & 20.58 & 12.82 & 16.24 & 14.31 & 16.78 & 25.14 & 23.56 & 12.81 & 79.04 & 56.83  \\
    CVT-Occ \cite{ye2024cvt} & 27.37 & 7.44 & 41.00 & 23.93 & 11.92 & 20.81 & 12.07 & 18.03 & 16.88 & \textbf{21.37} & 29.40 & 27.42 & 14.67 & 79.12 & 59.09 \\
    ALOcc-3D \cite{chen2025alocc} & 30.03 & 6.51 & 39.61 & 24.14 & 20.84 & 20.56 & 20.56 & 24.28 & 17.95 & 12.22 & 35.67 & 37.25 & 22.45 & 78.42 & 59.91 \\
    \midrule
    \rowcolor{pink!10} \textbf{Ours} & \textbf{31.25} & \textbf{9.03} & \textbf{42.74} & \textbf{27.52} & \textbf{22.91} & \textbf{25.38} & \textbf{21.47} & \textbf{27.84} & \textbf{22.65} & 20.69 & \textbf{39.98} & \textbf{38.89} & \textbf{24.02} & \textbf{81.21} & \textbf{64.46} \\
    \bottomrule
  \end{tabular}
  }
  \label{tab:waymo_results}
\end{table*}

\subsection{State-of-the-Art Comparison}
\label{sec:sota}

\vspace{0.5ex}\noindent \textbf{Results on Occ3D-nuScenes.}
We present a comprehensive quantitative comparison on the Occ3D-nuScenes validation set in Table \ref{tab:occ_cls}, which is organized into two settings: a single-frame setting without temporal voxel feature fusion (upper block) and a setting with one historical frame (\textbf{+1f}, lower block). Our framework achieves the best mIoU in both settings, reaching $41.2\%$ among single-frame methods and $42.2\%$ in the +1f setting, surpassing all competing methods under each respective protocol. By replacing strict ray constraints with FDR, our model gains consistent improvements on large-scale classes such as driveable surface, which require extensive cross-ray contextual reasoning. The Resolution-Context Decoupled Unification preserves local geometric fidelity, yielding competitive results on tightly localized objects such as \textit{car} and \textit{motor}.

\vspace{0.5ex}\noindent \textbf{Results on Occ3D-Waymo.}
The results on Occ3D-Waymo are summarized in Table \ref{tab:waymo_results}. Our method achieves a state-of-the-art mIoU of $\mathbf{31.25\%}$. Specifically, it outperforms ALOcc-3D \cite{chen2025alocc} by a margin of $\mathbf{+1.22\%}$ mIoU. The gains on spatially continuous classes, notably \textit{Road} ($81.21\%$) and \textit{Walkable} ($64.46\%$), confirm that FDR builds global contextual priors and reduces the spatial isolation inherent to pure physical-projection methods.

\subsection{Robustness in Uncalibrated Scenarios}
\label{sec:uncalibrated}

As theoretically analyzed in Section \ref{sec:method}, explicit physical projection methods are mathematically bound to strict geometric constraints. To empirically validate the resilience of FDR's global hypothesis space, we design an uncalibrated scenario on nuScenes. We completely withhold the dataset-provided camera parameters and instead rely on a lightweight convolutional head that blindly predicts rough camera poses from images. 
% Under such severe geometric perturbations, image features in classical physical baselines are erroneously splatted into incorrect spatial voxels, causing profound structural aliasing.
As shown in Table \ref{tab:uncalibrated}, explicit physical methods suffer catastrophic performance degradation. Our decoupled framework maintains a stable $28.0\%$ mIoU. This confirms that FDR's unconstrained global capacity lets the network internalize multi-camera rig topology from visual context alone, without relying on physical calibration.

\begin{table}[t]
    \centering
    \begin{minipage}[b]{0.33\textwidth}
        \caption{\textbf{Ablation on Model Components and Unification Strategies.} Evaluated under a lightweight single-frame setting on Occ3D-nuScenes.}
        \renewcommand{\arraystretch}{1.15} % Slightly looser for the taller table
        \resizebox{\textwidth}{!}{
        \begin{tabular}{l|c|c}
        \toprule
        \textbf{Component Strategy} & \textbf{mIoU} & \textbf{IoU} \\
        \midrule
        \textbf{Baseline (Our Default)} & \textbf{38.1} & \textbf{67.1} \\
        w/o Global Pathway (FDR) & 37.0 & 66.0 \\
        w/o Soft Geometric Hint & 37.7 & 66.7 \\
        Only BEV Representation & 37.7 & 67.9 \\
        Adaptive Weighting & 38.0 & 67.2 \\
        Late 3D Fusion & 37.7 & 66.9 \\
        \bottomrule
        \end{tabular}}
        \label{tab:ablation_component}
    \end{minipage}
    \hfill
    \begin{minipage}[b]{0.32\textwidth}
        \centering
        
        \caption{\textbf{Robustness without Camera Parameters.} Evaluated on nuScenes using implicitly predicted poses.}
        \renewcommand{\arraystretch}{1.2}
        \resizebox{\textwidth}{!}{
        \begin{tabular}{l|c|c}
        \toprule
        Method & \textbf{mIoU} & \textbf{IoU} \\
        \midrule
        BEVDetOcc \cite{huang2022bevdet4d} & 4.7 & 33.1 \\
        ALOcc \cite{chen2025alocc} & 9.0 & 43.5 \\
        FB-Occ \cite{li2023fbocc} & 12.0 & 54.5 \\
        \midrule
        \rowcolor{pink!10} \textbf{Ours} & \textbf{28.0} & \textbf{63.2} \\
        \bottomrule
        \end{tabular}}
        \label{tab:uncalibrated}
        \end{minipage}
    \hfill
    \begin{minipage}[b]{0.3\textwidth}
        
        \caption{\textbf{Ablation on Stage Configurations.} $V_t$ denotes volumetric expansion factor at each stage.}
        \renewcommand{\arraystretch}{1.8}
        \resizebox{\textwidth}{!}{
        \begin{tabular}{l|c|c}
        \toprule
        \textbf{Stage Schedule ($V_t$)} & \textbf{mIoU} & \textbf{IoU} \\
        \midrule
        \textbf{3 Stages (10$\to$5$\to$4)} & \textbf{38.1} & \textbf{67.1} \\
        \midrule
        2 Stages (20 $\to$ 10) & 37.4 & 66.5 \\
        3 Stages (4 $\to$ 5 $\to$ 10) & 37.7 & 66.8 \\
        3 Stages (10 $\to$ 10 $\to$ 2) & 37.9 & 66.9 \\
        \bottomrule
        \end{tabular}}
        \label{tab:ablation_stage}
        
    \end{minipage}
\end{table}

\subsection{Ablation Studies and Architectural Analysis}
\label{sec:ablation}

As shown in Tables \ref{tab:ablation_component} and \ref{tab:ablation_stage}, we conduct extensive ablation studies on the Occ3D-nuScenes dataset. For computational efficiency, all ablations are trained under a lightweight single-frame configuration with the channel dimension halved.

\vspace{0.5ex}\noindent\textbf{Model Design.} 
We first evaluate the necessity of the proposed dual-pathway unification in Table \ref{tab:ablation_component}. Removing the global pathway (FDR) entirely  degrades the performance of 1.1\%. This confirms that introducing fully-global contextual reachability is beneficial for overcoming the receptive field isolation of strict geometric priors. Additionally, omitting the soft geometric context ($\mathcal{E}_{geo}$) during the dynamic generation of FDR's routing weights yields a suboptimal $37.7\%$ mIoU, indicating that physical hints effectively guide the unconstrained dense mapping. We also observe that implementing the unification late in the network (\eg, after the 3D encoder, $37.7\%$), or substituting the tri-perspective representation with BEV representation ($37.7\%$) or adaptive spatial weighting ($38.0\%$), offers no discernible advantage. This affirms our design choice: because FDR and LSS occupy orthogonal structural regimes, simple element-wise addition on the gravity-aligned BEV plane is the natural and principled choice, avoiding additional engineering complexity.

\vspace{0.5ex}\noindent\textbf{Stage-wise Contraction Configurations.} 
We ablate the factorized spatial contraction across different $T$-stage settings in Table \ref{tab:ablation_stage}. We observe that compressing the dense routing into a shallower 2-stage network ($T=2$) degrades the mIoU to $37.4\%$, suggesting that an abrupt spatial aggregation hinders smooth contextual feature learning. Conversely, heavily delaying the volumetric expansion (\eg, 4-5-10 or 10-10-2 configurations) yields slightly lower results ($37.7\%$ and $37.9\%$) compared to our default progressive schedule (10-5-4, $38.1\%$). This shows that a progressive spatial abstraction schedule is crucial for building robust macroscopic context.
\begin{figure}[t]
    \centering
    \includegraphics[width=\textwidth]{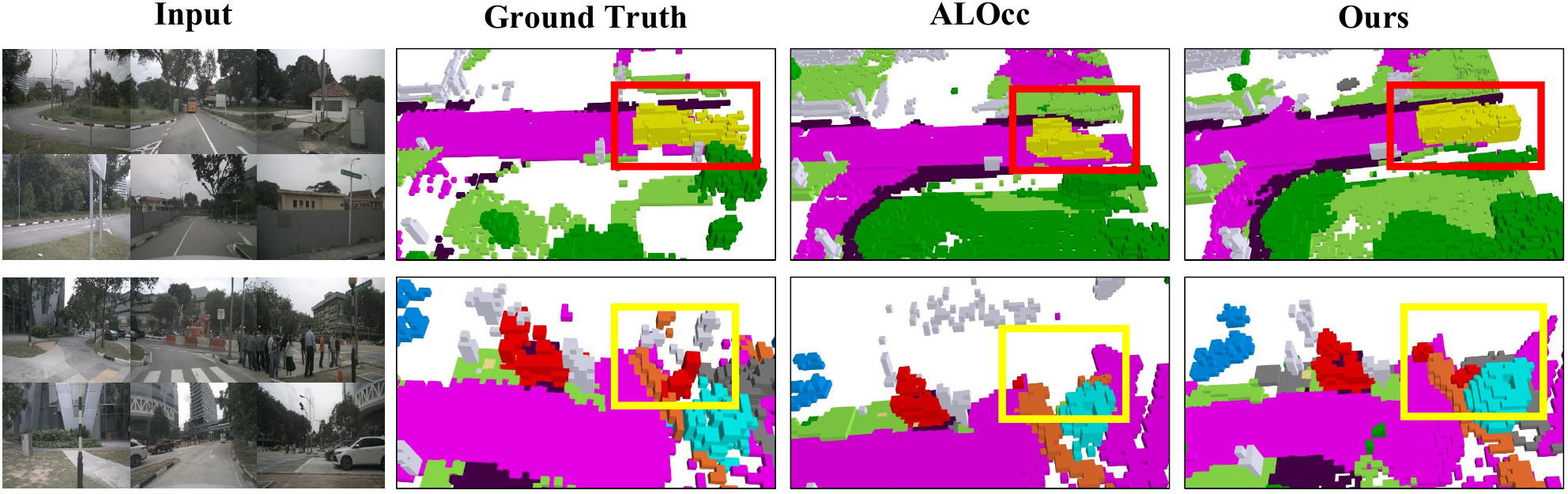}
    \caption{\textbf{Qualitative comparison of 3D semantic occupancy prediction.} We highlight the regions with complex structures and heavy occlusions in \textcolor{red}{red} and \textcolor{yellow}{yellow} boxes. While existing methods ALOcc often produce fragmented or incomplete occupancy topologies due to the strict ray-wise locality bottleneck, our proposed framework successfully recovers continuous and holistic structures by effectively internalizing global contextual relationships.}
    \label{fig:qualitative_results}
\end{figure}
\subsection{Qualitative Results}
Figure \ref{fig:qualitative_results} compares the occupancy predictions of our framework against ALOcc \cite{chen2025alocc} in complex scenes with heavy occlusions.
As shown, existing ray-based methods often produce fragmented or incomplete topologies, particularly for large objects spanning multiple cameras. This is a direct consequence of \textit{Receptive Field Isolation}: since individual 3D voxels can only aggregate evidence from strictly aligned rays, they fail to associate disjointed spatial clues into a coherent whole.
In contrast, our framework recovers continuous, holistic structures in occluded regions (highlighted in red and yellow boxes). By internalizing global context through \textit{Factorized Dense Routing}, our model establishes structural continuity where direct image evidence is absent. This empirical result verifies that our global routing operator effectively bridges the structural gaps that remain unreachable to sparse physical-masking baselines.

\section{Conclusion}
\label{sec:conclusion}

In this paper, we recast vision-based 3D occupancy prediction from rigid physical projection to generalized, unconstrained spatial routing. We identify the inherent \textit{Locality Bottleneck} of classical ray-based lifting and show that it is merely a structurally masked, degenerate subset of a fully-global hypothesis space. To harness this advantage, we propose \textbf{Factorized Dense Routing (FDR)}, scaling dense 2D-to-3D mapping to linear complexity via hierarchical spatial aggregations. Recognizing the inevitable \textit{Resolution-Context Trade-off} this factorization triggers, we architect a macro-micro decoupled framework that combines the global topological anchor from FDR with the pixel-exact precision of sparse physical priors.
Empirically, our framework establishes new state-of-the-art results across large-scale benchmarks. Most significantly, FDR lets the network internalize a fixed multi-camera rig purely from visual context, remaining robust even without physical calibration.

\vspace{0.5ex}\noindent\textbf{Limitations and Future Work.}
FDR attains a fully-global receptive field, but it still conditions its routing weights on a soft geometric hint from a lightweight depth head. This retains a residual dependence on explicit geometry, and a fully geometry-free variant remains an open direction. Moreover, our uncalibrated study uses a single fixed rig shared across scenes. Cross-rig, calibration-free generalization is thus a natural next step toward robust, geometry-agnostic 3D representation learning.

\clearpage  % TODO FINAL: This \clearpage needs to be removed from both review and camera-ready versions.

% \section*{Acknowledgements}
% Please insert your acknowledgments here.

\bibliographystyle{splncs04}
\bibliography{main}
\clearpage
\appendix
\renewcommand{\thesection}{\Alph{section}}
\renewcommand{\thetable}{\Alph{section}\arabic{table}}
\renewcommand{\thefigure}{\Alph{section}\arabic{figure}}

\section{Derivation of the Computational Reduction Ratio}
\label{supp:complexity}

In Section 3.3 of the main text, we state that the proposed Factorized Dense Routing (FDR) circumvents the quadratic bottleneck of global dense routing, requiring only $\sim 1.4\%$ of its theoretical computation under practical settings. Here we provide the stage-wise mathematical derivation.

\noindent\textbf{Exact Formulation of the Complexity.} 
Explicitly calculating the dense routing matrix between all 2D perspective pixels $S_0$ and all 3D ego-voxels $V_T$ incurs a prohibitive computational complexity: $\mathcal{C}_{global} = \mathcal{O}(S_0 \cdot V_T \cdot C)$.

In contrast, FDR factorizes this dense mapping into $T$ localized stages. At any stage $t \in [1, T]$, the routing is bounded within a contracted 2D spatial size $S_t$ and a progressively expanding 3D volume size $V_{t-1}$. Given the adaptive 2D patch size $P_t$ and the 3D micro-volume expansion factor $K_t$, the complexity of a single FDR stage is:
\begin{equation}
    \mathcal{C}^{(t)}_{FDR} = \mathcal{O}\Big( S_t \cdot V_{t-1} \cdot P_t \cdot K_t \cdot C \Big) 
\end{equation}
Because $S_{t-1} = S_t \cdot P_t$ and $V_t = V_{t-1} \cdot K_t$, the stage-wise complexity reduces to $\mathcal{C}^{(t)}_{FDR} = \mathcal{O}( S_{t-1} \cdot V_t \cdot C )$.

We define the theoretical reduction ratio $\eta_t = \mathcal{C}^{(t)}_{FDR} / \mathcal{C}_{global}$. By separating the spatial domains, the fraction of computation at stage $t$ simplifies to:
\begin{equation}
    \eta_t = \frac{S_{t-1}}{S_0} \times \frac{V_t}{V_T} = \left( \prod_{j=1}^{t-1} P_j \right)^{-1} \left( \prod_{j=t+1}^{T} K_j \right)^{-1}
    \label{eq:eta_t}
\end{equation}
The total relative complexity is the sum across all $T$ stages: $\eta = \sum_{t=1}^T \eta_t$. 

\noindent\textbf{Practical Case Analysis: The $\sim 1.4\%$ Overhead.} 
In our default configuration ($T=3$), the 2D spatial grid is progressively contracted by patch areas of $P_1=16$ ($4\times4$), $P_2=8$ ($2\times4$), and $P_3=4$ ($2\times2$). Conversely, since we compress the Z-axis to $1$ during FDR, the micro-volume expansion operates strictly on the X-Y BEV plane, with per-stage area factors of $K_1=100$ ($10\times10$), $K_2=25$ ($5\times5$), and $K_3=16$ ($4\times4$). These factors fully expand the BEV plane to the target $200\times200$ resolution (\ie, $\prod_t K_t = 4\times10^4$).

According to Eq.~\eqref{eq:eta_t}, the total complexity ratio for $T=3$ expands to:
\begin{equation}
    \eta = \underbrace{\frac{1}{K_2 K_3}}_{\text{Stage 1}} + \underbrace{\frac{1}{P_1 K_3}}_{\text{Stage 2}} + \underbrace{\frac{1}{P_1 P_2}}_{\text{Stage 3}}
\end{equation}
Substituting our practical network constants yields:
\begin{equation}
    \begin{aligned}
    \eta &= \frac{1}{K_2 K_3} + \frac{1}{P_1 K_3} + \frac{1}{P_1 P_2}
    = \frac{1}{25 \times 16} + \frac{1}{16 \times 16} + \frac{1}{16 \times 8} \\
    &= \frac{1}{400} + \frac{1}{256} + \frac{1}{128} \approx 0.0142
    \end{aligned}
\end{equation}
As shown, the early stages are kept cheap by the still-small 3D volume ($\eta_1\!=\!1/400$), while the late stages are kept cheap by the heavily contracted 2D grid ($\eta_3\!=\!1/128$). The combined cost $\eta$ is thus roughly $\mathbf{1.42\%}$ of the naive dense routing equivalent, confirming the efficiency of FDR claimed in the main text.

\noindent\textbf{Generalized Asymptotic Bound.} 
Beyond the specific case of $T=3$, the efficiency of FDR holds for any network depth $T$. 

Let $M = \min(\{P_1 \dots P_T\}, \{K_1 \dots K_T\})$ denote the minimum spatial contraction or volumetric expansion rate across the entire network. By observing Eq.~\eqref{eq:eta_t}, the denominator of each term $\eta_t$ is the product of exactly $(t-1) + (T-t) = T-1$ sequential scaling variables. Thus, we can establish a strict upper bound for the total complexity ratio:
\begin{equation}
    \eta = \sum_{t=1}^T \eta_t \le \sum_{t=1}^T \frac{1}{M^{T-1}} = \frac{T}{M^{T-1}}
\end{equation}
This bound, $\mathcal{O}(T / M^{T-1})$, gives a worst-case guarantee on the relative overhead of factorized routing. For fixed $T$, the overhead falls quickly as $M$ grows. The bound also certifies that factorized routing avoids the $\mathcal{O}(N^2)$ cost of naive global routing at large spatial resolutions, confirming its scalability to high-resolution inputs.

\section{Proof that LSS is a Degenerate Special Case of FDR}
\label{supp:proof}

In Section 3.3 of the main text, we claimed that classical explicit physical projection (LSS~\cite{philion2020lift,chen2025alocc}) is a degenerate special case of FDR, \ie, $\mathcal{F}_{LSS} \subseteq \mathcal{F}_{FDR}$. Here we give the full derivation, including tensor alignments and the required approximation conditions.

\noindent\textbf{Step 1: The LSS Matrix Formulation.} 
Given camera calibration $\Theta$, LSS predicts a categorical depth distribution $\mathcal{D} \in \mathbb{R}^{S_0 \times D}$ along the physical ray of each 2D pixel $s$. We formulate its physical geometry via a sparse binary indicator tensor $\mathbf{M}_\Theta \in \{0, 1\}^{S_0 \times D \times V_T}$, where $\mathbf{M}_\Theta(s, d, v) = 1$ if and only if the $d$-th depth bin of pixel $s$'s ray falls precisely into the 3D ego-voxel $v$. The LSS feature aggregation is defined as:
\begin{equation}
    \mathcal{V}_{LSS}(v) = \sum_{s=1}^{S_0} \sum_{d=1}^D \mathcal{F}_{2D}(s) \cdot \mathcal{D}(s, d) \cdot \mathbf{M}_\Theta(s, d, v)
    \label{eq:lss_agg}
\end{equation}

\noindent\textbf{Step 2: Degenerate FDR Instantiation.}
We instantiate FDR with a single stage ($T=1$). To eliminate cross-patch macroscopic aggregation, we enforce no spatial contraction by setting $P_1 = 1 \times 1$. Thus, the $\text{Unfold}_{P_1}$ operator degrades to an identity mapping, preserving the original 2D resolution ($S_1 = S_0$). We map the micro-volume expansion strictly to the depth dimension by setting $K_1 = D$. 

Under these hyperparameters, the FDR local dense routing simplifies to a pure depth-wise element multiplication along each uncontracted pixel:
\begin{equation}
    \mathcal{Y}^{(1)}(s, d) = \mathcal{F}_{2D}(s) \cdot \mathbf{A}^{(1)}(s, d)
    \label{eq:degen_route}
\end{equation}
where $\mathbf{A}^{(1)} \in \mathbb{R}^{S_0 \times D}$ is the dynamically predicted routing probability matrix. 

\noindent\textbf{Step 3: Weight Construction and Masking.}
In FDR, $\mathbf{A}^{(1)}$ is generated by the lightweight network $g_\theta(\mathcal{F}_{2D}, \mathcal{E}_{geo})$ followed by a Softmax $\sigma$. Since the construction only requires the existence of one feasible parameterization, we directly assign the routing logits to equal the LSS depth logits, \ie, we choose weights $\theta$ such that $\sigma(g_\theta(\cdot)) = \mathcal{D}$, which is realizable because $\mathcal{D}$ is itself the output of an analogous depth-prediction head. The degenerate special case additionally imposes the geometric mask $\mathbf{M}_\Theta$ prior to folding, enforcing the same ray-wise structural sparsity that LSS assumes. Thus, $\mathbf{A}^{(1)}(s, d) \equiv \mathcal{D}(s, d)$.

\noindent\textbf{Step 4: Fold Equivalence.}
The final $\text{Fold}$ operation in FDR projects the expanded $(s, d)$ space back into the global $V_T$ space. By applying the exact spatial correspondence defined by $\mathbf{M}_\Theta$, the Fold operator executes the following summation:
\begin{equation}
    \mathcal{V}_{FDR}(v) = \sum_{s=1}^{S_0} \sum_{d=1}^D \mathcal{Y}^{(1)}(s, d) \cdot \mathbf{M}_\Theta(s, d, v)
    \label{eq:fdr_fold}
\end{equation}
Substituting Eq.~\eqref{eq:degen_route} into Eq.~\eqref{eq:fdr_fold}, with $\mathbf{A}^{(1)} \equiv \mathcal{D}$ from Step 3, recovers the LSS aggregation in Eq.~\eqref{eq:lss_agg}.

\textbf{Conclusion:} Any view transformation executed by LSS can be reproduced by a single-stage FDR with $P_1=1$ and the geometric mask $\mathbf{M}_\Theta$. Therefore $\mathcal{F}_{LSS} \subseteq \mathcal{F}_{FDR}$. \hfill $\blacksquare$

\section{Extended Implementation Details}
\label{supp:implementation}

\noindent\textbf{Network Architecture.} 
The 2D image backbone utilizes a pre-trained ResNet-50 equipped with a Feature Pyramid Network (FPN), which projects multi-scale perspective features into a unified channel dimension of $C=80$. For the dynamic weight generation network $g_\theta$, we use a lightweight block of two consecutive $3\times3$ convolutional layers (BatchNorm and ReLU), followed by a linear projection to produce the $K_t$ routing logits. 
The explicit geometric context $\mathcal{E}_{geo}$ is formed by concatenating lightweight depth priors with Plücker ray embeddings. The Plücker coordinates are mapped to the high-dimensional space via a 2-layer MLP with a hidden dimension of 64. 
The 2D convolutional encoders processing the unified BEV plane $\tilde{\mathcal{P}}_{xy}$ and the geometric planes $\mathcal{P}_{xz}, \mathcal{P}_{yz}$ consist of standard ResNet basic blocks, maintaining the $C=128$ dimension before interacting with the final MLP classification head.

\noindent\textbf{Uncalibrated Robustness Evaluation Setup.} 
In Section 4.3 of the main text, we evaluated our framework's robustness when ground-truth camera extrinsics are entirely withheld. In the absence of explicit matrices, the network inherently loses the spatial ordering of the multi-view cameras. To address this, we inject view-specific learnable embeddings into the 2D features to encode Camera ID. We then attach a lightweight CNN head (three $3\times3$ convolutional layers, Global Average Pooling, and a linear regressor) that is jointly optimized with the backbone to predict 6-DoF relative camera extrinsics from the view-aware features. The resulting noisy pose estimates are used for view transformation. To fairly evaluate raw routing robustness, all baselines (BEVDetOcc, FB-Occ, ALOcc) and our model are retrained from scratch with the same predicted poses and view embeddings.

\section{Empirical Efficiency Analysis}
\label{supp:efficiency}
\begin{table}[t]
  \centering
  \caption{\textbf{Empirical Inference Speed.} Evaluated on Occ3D-nuScenes using a single NVIDIA A100 GPU. Our framework achieves state-of-the-art mIoU while maintaining competitive real-time efficiency.}
  \renewcommand{\arraystretch}{1.1}
  \resizebox{0.32\columnwidth}{!}{
  \begin{tabular}{l|c|c}
   \toprule
    Method & \textbf{FPS} & \textbf{mIoU} \\
    \midrule
    BEVDetOcc \cite{huang2022bevdet4d} & 8.2 & 37.1 \\
    ALOcc \cite{chen2025alocc} & 7.8 & 40.1 \\
    \midrule
    \rowcolor{pink!10} \textbf{Ours} & 7.4 & \textbf{41.2} \\
    \bottomrule
  \end{tabular}
  }
  \label{tab:supp_efficiency}
\end{table}

We further evaluate the empirical inference speed (FPS) to ensure practical deployability. Measurements are conducted on a single NVIDIA A100 GPU using the Occ3D-nuScenes single-frame setup (ResNet-50 backbone, $256 \times 704$ resolution).
As shown in Table \ref{tab:supp_efficiency}, despite a dual-pathway architecture capturing both global context and full spatial resolution, our framework runs at $7.4$ FPS. This speed closely matches the explicit projection baseline ALOcc \cite{chen2025alocc}, indicating that the $\text{Unfold} \to \text{MatMul} \to \text{Fold}$ contractions in FDR map efficiently onto modern GPU hardware.

\end{document}